# Control Neuronal por Modelo Inverso de un Servosistema Usando Algoritmos de Aprendizaje Levenberg-Marquardt y Bayesiano.

**Victor A. Rodriguez-Toro[§], Jaime E. Garzón[‡], Jesús A. López[†]**

[§]*Grupo de Investigación en Bionanoelectrónica. Escuela de Ingeniería Eléctrica y Electrónica.
A.A. 25360, Universidad del Valle, Cali, Colombia (email: victor.rodriguez@correounivalle.edu.co)*

[‡] *Escuela de Ingeniería Eléctrica y Electrónica.
A.A. 25360, Universidad del Valle, Cali, Colombia (email: jaimeegv@hotmail.com)*

[†]*Grupo de Investigación en Energías GIEN-UAO. Departamento de Automática y Electrónica.
Calle 25 No. 115-85, Universidad Autónoma de Occidente, Cali, Colombia (e-mail: jalopez@uao.edu.co)*

**Resumen:** En este trabajo se presentan los resultados experimentales del control neuronal de velocidad de un servosistema. La estrategia de control neuronal implementada es el control por modelo inverso. El entrenamiento de la red se ha realizado usando dos algoritmos de aprendizaje: Levenberg-Marquardt y regularización Bayesiana. Se evalúa la capacidad de generalización de cada método en función del correcto funcionamiento del controlador para seguir la señal de referencia y de los esfuerzos de control obtenidos

*Palabras Clave:* Neuro-Control, Red Neuronal Artificial, Algoritmos de Entrenamiento, Control por Dinámica Inversa

## 1. INTRODUCCIÓN

Una de las propiedades más explotadas de las redes neuronales artificiales (RNA) es la de ser aproximador universal de funciones [Hornik 89], lo cual ha llevado a utilizarlas en diversidad de aplicaciones que van desde el reconocimiento de patrones [Bishop 95], hasta la identificación y/o control de sistemas dinámicos [Narendra 90].

En el área de control de sistemas dinámicos, las RNA se han usado en diferentes enfoques siendo el control neuronal por modelo inverso la estrategia básica de control neuronal [Norgaard 00].

Además, debe señalarse que a pesar de existir una gran diversidad de algoritmos de aprendizaje para las redes neuronales, los más usados para tal fin son los algoritmos basados en gradiente. Dependiendo de la manera como se utilice la información del gradiente, se tienen algoritmos de aprendizaje denominados de primer orden como el gradiente descendente [Hagan 96, Haykin 99] o algoritmos de aprendizaje más elaborados denominados de segundo orden como los algoritmos de gradiente conjugado [Hagan 96, Haykin 99] y los algoritmos basados en la metodología de Levenberg-Marquardt [Hagan 96, Masters 95].

En aplicaciones de control neuronal por modelo inverso, generalmente se selecciona como algoritmo de aprendizaje el basado en Levenberg-Marquardt, debido principalmente a su rapidez de convergencia con errores suficientemente pequeños acorde a las necesidades del usuario. Sin embargo, alcanzar errores pequeños no garantiza la capacidad de generalización del modelo neuronal, produciendo en este caso un problema de sobre-entrenamiento, lo cual significa que la red presenta un error muy bajo con los patrones de entrenamiento pero el error aumenta con los patrones de validación.

Entre las diferentes alternativas para evitar el sobre-entrenamiento se destaca el algoritmo de regularización automática o aprendizaje Bayesiano. En aplicaciones de identificación de sistemas, esto se ha explorado teniendo como resultado que la calidad de los modelos neuronales mejora cuando se usa aprendizaje Bayesiano en vez de aprendizaje convencional como Levenberg-Marquardt [Lopez 07]

El objetivo de este trabajo es ver las diferencias entre controladores neuronales entrenados con métodos que no garantizan la capacidad de generalización como el Levenberg-Marquardt y métodos que si consideran este aspecto como el aprendizaje Bayesiano Este documento está organizado de la siguiente manera: En la sección 2 se presentan conceptos generales de control neuronal y de entrenamiento de redes neuronales considerando y no el sobre-entrenamiento. Posteriormente, en la sección 3, se presenta la implementación del control neuronal por modelo inverso en el servosistema. En la sección 4 se muestran los resultados experimentales. El trabajo termina con la sección 5 donde se presentan las conclusiones.

## 2. CONTROL POR MODELO INVERSO: APRENDIZAJE LEVENBERG-MARQUARDT Y BAYESIANO.

*2.1 Modelo inverso*

El control por modelo inverso es una técnica, que busca cancelar la dinámica de la planta al colocar un elemento en cascada con ella, en este caso una red neuronal, siendo este una aproximación matemática del inverso de la planta. De esta manera se busca que la salida sea lo más parecida posible a la referencia [Norgaard 00].

Existen dos tendencias cuando se realiza control neuronal por modelo inverso, la primera es la conocida como entrenamiento general en el cual la red neuronal usando datos obtenidos con anterioridad encuentra el modelo inverso de la planta. Cuando la red neuronal ha sido entrenada se puede usar como controlador pues ella cancela la dinámica de la planta como se observa en la Fig. 1

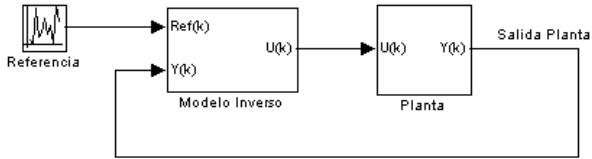

Fig. 1. Esquema de Control por Modelo Inverso

La otra tendencia, conocida como entrenamiento especializado, es un esquema de control adaptativo donde el objetivo es minimizar el error entre la salida de la planta (y(k)) y la salida del modelo de referencia ($y_m(k)$) [Norgaard 00]. Para este entrenamiento se requiere otra red neuronal (ya entrenada) que sea el modelo directo de la planta (Fig. 2).

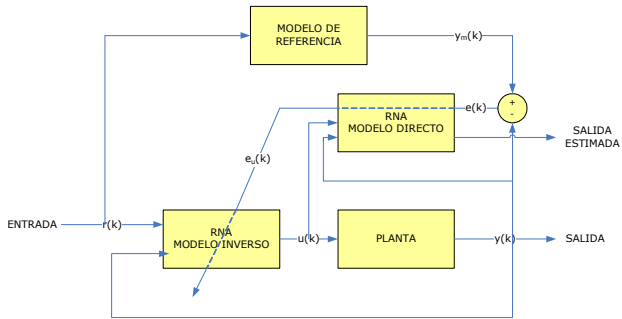

Fig. 2. Estructura para entrenamiento inverso especializado.

*2.2 Algoritmo de Aprendizaje Levenberg-Marquardt (LM)*

En la ecuación (1), se presenta cómo se localiza un valor mínimo ($x_{min}$) de una función de una variable f(x), utilizando la primera y segunda derivada de acuerdo al método de Newton.

$$x_{min}(t+1) = x_{min}(t) - \frac{f'(x_{min}(t))}{f''(x_{min}(t))} \quad (1)$$

Con base en esta ecuación se puede inferir (2), donde se minimiza el error global $E_p$ en el espacio de los pesos sinápticos representado por la matriz **W**.

$$\mathbf{W}(t+1) = \mathbf{W}(t) - \frac{E_P'}{E_P''} \quad (2)$$

La segunda derivada del error global ($E_P''$) corresponde a la Matriz Hessiana **H** y la primera derivada ($E_P'$) la conocemos como el vector gradiente **G**. El vector gradiente y la matriz Hessiana de la función de error los podemos calcular utilizando la regla de la cadena. Así, el vector gradiente está compuesto por las derivadas parciales del error con respecto a cada uno de los pesos $w_i$ de la red, el elemento (i,j) de la matriz Hessiana se calcula con las segundas derivadas parciales del error con respecto a los pesos $w_i$ y $w_j$.

Debido a la carga computacional que implica calcular de manera exacta la matriz H, se hace una estimación de la misma [Master 95]. Debido a esto, en (2) se introduce un mecanismo de control para evitar los problemas que se puedan tener en la actualización de pesos de la red, dando origen a (3)

$$\mathbf{W}(t+1) = \mathbf{W}(t) - (\mathbf{H} + \lambda\mathbf{I})^{-1}\mathbf{G} \quad (3)$$

El mecanismo de control para garantizar la convergencia del algoritmo consiste en introducir un factor $\lambda\mathbf{I}$. En primer lugar se prueba la ecuación del método de Newton. Si al evaluarla, el algoritmo no converge (el valor del error comienza a crecer), se elimina este valor y se incrementa el valor de $\lambda$ en (3), con el fin de minimizar el efecto de la matriz **H** en la actualización de los pesos. Si $\lambda$ es muy grande, el efecto de la matriz **H** prácticamente desaparece y la actualización de pesos se hace esencialmente con el algoritmo de gradiente descendente. Si el algoritmo tiene una clara tendencia hacia la convergencia se disminuye el valor de $\lambda$ con el fin de aumentar el efecto de la matriz **H** y de esta manera se garantiza que el algoritmo se comporta con un predominio del Método de Newton.

El método Levenberg Marquardt mezcla sutilmente el método de Newton y el método Gradiente Descendente en una única ecuación para estimar la actualización de los pesos de la red neuronal.

*2.3 Regularización y algoritmo de Aprendizaje por Regularización Bayesiana (BR)*

En algunas aplicaciones se observa que las redes neuronales artificiales pueden caer en un problema conocido como sobre-entrenamiento [Bishop 95], en donde la red no es capaz de responder adecuadamente ante datos de entrada diferentes a los datos que se utilizaron para el proceso de aprendizaje, pero que hacen parte del problema que se quiere solucionar. Visto de otra manera, la red se especializa o "memoriza" un conjunto de datos determinados con un error muy pequeño.

Este sobre-entrenamiento trae como consecuencia que el error de test o verificación de la red sea mucho mayor que el de entrenamiento lo cual es indeseable cuando la red está trabajando en la solución de una aplicación específica.

Surge entonces esta técnica, denominada regularización, cuyo objetivo es minimizar el fenómeno del sobre-entrenamiento y por ende sus efectos. El fenómeno del sobre-entrenamiento se hace más evidente cuando los datos de entrada están contaminados con ruido. El objetivo de la regularización es garantizar un adecuado aprendizaje evitando este problema.

*2.3.1 Regularización por Parada Temprana*

En el proceso de aprendizaje se definen dos tipos de error: el primero es el de aprendizaje, cuya evolución la mostramos en la figura 3 con la línea continua y se calcula como la diferencia entre la salida de la red y el valor deseado. Este error es el que se utiliza para modificar los pesos de la red.

Con esta técnica, surge ahora, un segundo error que se denomina de validación, cuya evolución se muestra con la línea punteada y se calcula de igual manera, como la diferencia entre la salida de la red y el valor deseado, pero evaluado en un dato de entrada que pertenece al conjunto de datos de validación. Es importante aclarar que este error no se considera para la modificación los pesos sinápticos de la red, por lo que en este momento se observa el comportamiento de la red frente a nuevos datos que pertenecen al conjunto universo del problema, pero diferentes a los utilizados en el entrenamiento.

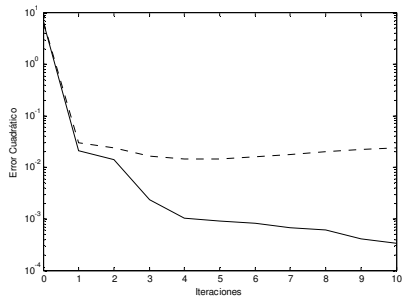

Fig. 3 Regularización por Parada Temprana. Línea Continua: Error de entrenamiento; Línea Punteada: Error de validación

En la Fig. 3 se observa como el error de entrenamiento puede seguir decreciendo a pesar de que el error de validación comenzó a subir. Esto es una señal de sobre-entrenamiento y para evitarlo, se recomienda detener el proceso de entrenamiento de la red justo antes de que el error de validación empiece con una tendencia ascendente. Por esta razón, se llama "Regularización por Parada Temprana" [Bishop 95], debido a que se evita que el proceso de aprendizaje continúe disminuyendo excesivamente el error de entrenamiento, tomando como criterio de finalización la tendencia ascendente del error de validación.

*2.3.2 Regularización por Limitación de la Magnitud de los Pesos*

La experiencia ha mostrado que se puede garantizar una función de salida suave si se logra mantener los pesos sinápticos de la red en unos valores relativamente pequeños [Bishop 95]. Con el fin de limitar la magnitud de estos pesos se redefine el cálculo del error así:

$$E_R = \alpha E_D + \beta E_W \qquad (4)$$

En (4) el error se calcula con dos términos, por lo que se utiliza una nueva notación y por lo tanto, se denominará a $E_R$ al Error Regularizado. El primer término, corresponde al error cuadrático promedio, presentado en (5), que es la forma como tradicionalmente se ha estimado el error de entrenamiento, pero afectado por el parámetro α.

$$E_D = \frac{1}{2P}\sum_{p=1}^{P} \sum_{k=1}^{M}(d_{pk} - y_{pk})^2 \qquad (5)$$

El segundo término utilizado para estimar el error de aprendizaje regularizado, introduce la sumatoria de los pesos sinápticos de la red, mostrado en (6), ponderados con el parámetro β. Debido a que el algoritmo de aprendizaje tiende a minimizar $E_R$, el resultado final es que la sumatoria de pesos igualmente tiende a ser minimizada para garantizar una suavidad en la salida de la red.

$$E_W = \sum_{n=1}^{N} w_n^2 \qquad (6)$$

*2.3.3 Regularización Bayesiana*

Trabajar con la técnica del error regularizado mostrado en (4) tiene el inconveniente que hay que definir los parámetros α y β de tal manera que se consiga el efecto de minimizar el sobre-entrenamiento [Lopez 05]. La regularización Bayesiana es una técnica donde entrenamiento de la red neuronal se aborda desde una perspectiva probabilística donde se considera una distribución de probabilidad en los valores de los pesos [Mackay 96]. A esta técnica de regularización también se le puede denominar regularización automática pues se trabaja con el error regularizado pero de tal forma que el algoritmo de aprendizaje encuentra los valores más adecuados para los parámetros regularizantes α y β [Foresee 97]

## 3. CONTROL NEURONAL POR MODELO INVERSO DEL MOTOR DC

*3.1 Servosistema: Motor DC*

Como proceso a controlar se utilizó un motor de DC, cuyo eje está acoplado a un disco y en el cual se pretende controlar la velocidad de rotación. En esta planta los datos son enviados desde el PC hacia el motor a través de una tarjeta de adecuación, mientras que existe un registro directo y permanente sobre la velocidad obtenida por medio de un tacómetro. En la Fig. 4 se observa una fotografía del servosistema a controlar

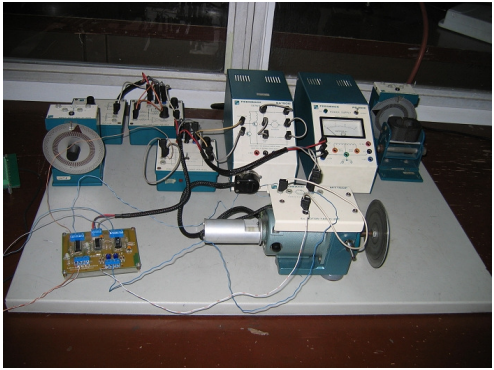

Fig. 4. Servomecanismo a controlar.

El servosistema posee una zona muerta de aproximadamente 2.5V, y la zona de saturación detectada a partir de los 6.5V. Este par de valores se convierten en cotas superior e inferior para el proceso de identificación y entrenamiento de la red neuronal, y por tanto, la acción de control operará de la forma deseada dentro de este rango de trabajo.

3.2 *Control Neuronal Por Modelo Inverso Del Motor DC*

La idea de realizar control por medio de técnicas neuronales fue inicialmente planteada en [Narenda 90]. El objetivo es aprovechar la capacidad de aprendizaje de las redes neuronales para entrenar redes que puedan ser usadas como controladores.

Inicialmente se realiza el Control por Modelo Inverso [Norgaard 00]. El modelo inverso en este caso de estudio, es una red neuronal convenientemente entrenada para que haga la labor de controlador.

*3.2.1 Arquitectura interna de la red neuronal.*

La red neuronal que realiza el control es una red del tipo Perceptron Multicapa con una capa oculta. La red tiene 5 entradas que son: el valor de referencia, la muestra actual de la salida del sistema, la muestra anterior de la salida del sistema, la muestra anterior de la acción de control y la acción de control con dos retardos *(ref(k),y(k),y(k-1),u(k-1),u(k-2))*, con esta información la red estimará la acción de control a aplicar al proceso *(u(k))*. La capa oculta está conformada por un total de diez neuronas con funciones de activación tangente-sigmoidales.

3.3 *Descripción Del Experimento.*

Antes de definir el experimento, se realizó una prueba en estado estable para identificar la zona muerta y la zona de saturación para determinar la región de operación del motor como se mencionó en la sección 3.1 estos valores están entre los 2.5V y los 6.5V, respectivamente

*3.3.1 Toma de datos*

Para entrenar un controlador neuronal por modelo inverso, lo primero que hay que hacer es realizar un experimento para tomar datos del funcionamiento del motor en su región de operación.

En este experimento se aplican escalones con amplitud aleatoria al motor la duración de escalón es tal que se garantice la estabilización de la velocidad del motor. Usando tarjetas de adquisición de datos se capturan las señales de entrada y la salida de velocidad del servomecanismo. Para la realización de este experimento se uso la herramienta LabView

*3.3.2 Entrenamiento de la red neuronal*

El esquema de control neuronal utilizado fue el control neuronal inverso general. En este caso el modelo inverso se entrenó off-line usando para tal fin el toolbox de redes neuronales que tiene MATLAB.

Para el entrenamiento de la red neuronal se utilizó como métodos de entrenamiento el aprendizaje basado en Levenberg Marquardt y el aprendizaje con Regularización Bayesiana.

## 4. RESULTADOS EXPERIMENTALES

Luego de entrenar las redes neuronales con las estrategias de entrenamiento mencionadas, se procede a verificar el comportamiento de los neuro-controladores obtenidos. Para hacer esta verificación se implementó una aplicación en la herramienta LabView donde se toman los pesos de las redes neuronales entrenadas en MATLAB y se usan como neuro-controladores inversos usando un esquema similar a la de la figura 1.

Vale la pena mencionar que en todas las redes probadas como neuro-controladores se obtuvieron errores de entrenamiento similares, de tal manera que lo que se verifica cuando se prueban al controlar la planta es la capacidad de generalización de las mismas.

*4.1 Controles inversos entrenados con Levenberg Marquardt.*

Los primeros neuro-controladores que se usaron en el servosistema fueron los obtenidos con el entrenamiento basado en la metodología de Levenberg Marquardt.

En las figuras 5 y 6 se observa el comportamiento típico de uno de estos neuro-controladores. Como se observa, el funcionamiento no es el más adecuado pues no se logra el seguimiento de la señal de referencia y los esfuerzos de control son muy oscilatorios.

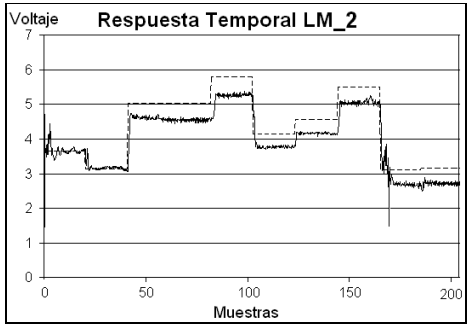

Fig. 5. Respuesta temporal del neuro-controlador entrenado con Levenberg Marquardt. Línea punteada: Referencia; Línea definida: Salida de la planta.

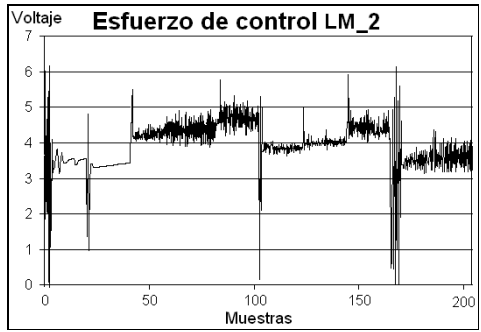

Fig. 6. Respuesta temporal para el esfuerzo de control del neuro-controlador entrenado con Levenberg Marquardt.

Sin embargo, se realizaron varias pruebas de entrenamiento con esta metodología hasta encontrar una donde se cumplieran de manera satisfactoria el control del servosistema. En las figuras 7 y 8 se observa el mejor neuro-controlador obtenido, aunque se logra un mejor seguimiento de la señal de referencia que el caso anterior, los esfuerzos de control siguen siendo bastante oscilatorios.

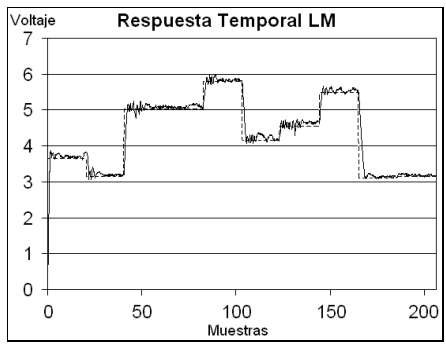

Fig. 7. Respuesta temporal del neuro-controlador entrenado con Levenberg Marquardt. Línea punteada: Referencia; Línea definida: Salida de la Planta.

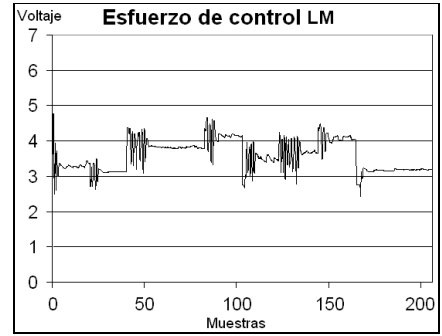

Fig. 8. Respuesta temporal para el esfuerzo de control del neuro-controlador entrenado con Levenberg Marquardt.

### 4.2 Controles inversos entrenados con Regularización Bayesiana.

Posteriormente se entrenaron neuro-controladores inversos usando el aprendizaje con regularización Bayesiana. En las figuras 9 y 10 se observa el comportamiento típico de uno de estos neuro-controladores, se puede observar que el seguimiento de la señal de referencia es mucho mejor que los presentados en la sección 4.1 además, el esfuerzo de control no presenta oscilaciones tan notorias.

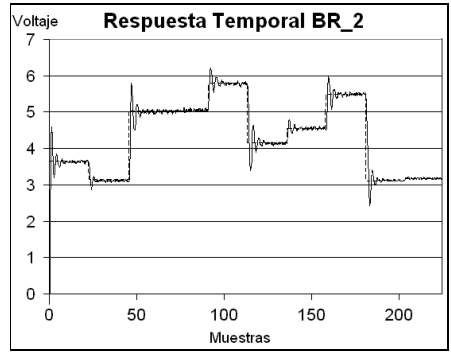

Fig. 9. Respuesta temporal del neuro-controlador entrenado con regularización Bayesiana. Línea punteada: Referencia; Línea definida: Salida de la Planta.

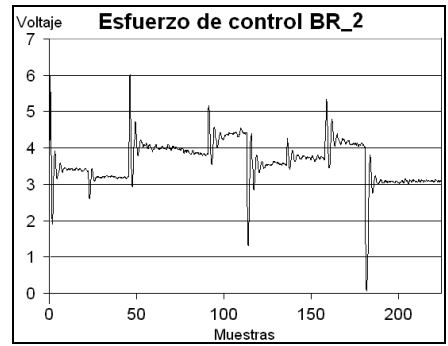

Fig. 10. Respuesta temporal para el esfuerzo de control del neuro-controlador entrenado con regularización Bayesiana.

En las figuras 11 y 12 se observa el comportamiento de uno de los neuro-controladores entrenados con regularización Bayesiana que mejor funcionó. Obsérvese el seguimiento de la referencia con pocos sobre impulso y los esfuerzos de control con oscilaciones de menor amplitud.

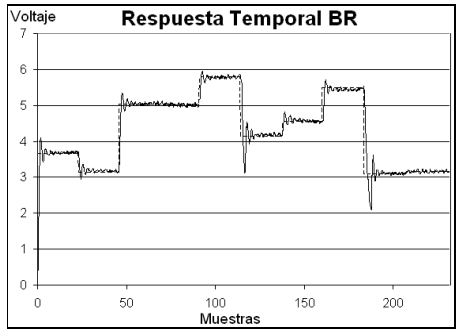

Fig. 11. Respuesta temporal del neuro-controlador entrenado con regularización Bayesiana. Línea punteada: Referencia; Línea definida: Salida de la Planta.

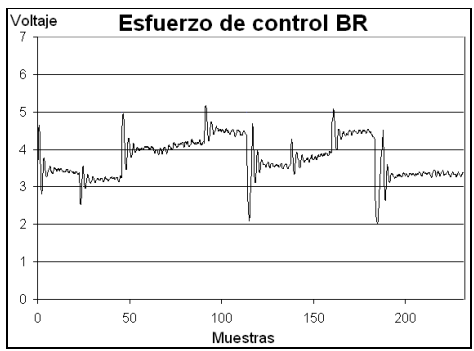

Fig. 12. Respuesta temporal para el esfuerzo de control del neuro-controlador entrenado con regularización Bayesiana.

*4.3 Índices de desempeño.*

Para estimar cualitativamente el desempeño de los controladores entrenados con Levenberg Marquardt y regularización Bayesiana; se calcularon como índices de desempeño el promedio del error absoluto y el promedio del esfuerzo de control.

**Tabla I. Índices de desempeño para el controlador con Levenberg Marquardt y regularización Bayesiana.**

|  | Levenberg Marquardt | Regularización bayesiana |
|---|---|---|
| Promedio del error absoluto | 0,2434632 | 0,08789647 |
| Índice de esfuerzo de control | 3,81427187 | 3,74021221 |

Al analizar los índices de desempeño obtenidos, se puede concluir que el seguimiento del neuro-controlador entrenado con regularización Bayesiana es mejor que el logrado con Levenberg Marquardt pues tiene un menor valor del promedio del error absoluto. Una situación similar se tiene con el esfuerzo de control, pues el neuro-controlador entrenado con regularización Bayesiana presenta un menor valor de este índice de desempeño.

## 5. CONCLUSIONES

Tener en cuenta el efecto del sobre-entrenamiento en el entrenamiento de neuro-controladores inversos puede mejorar el desempeño del mismo.

El mejor desempeño de los neuro-controladores entrenados con aprendizaje Bayesiano se puede explicar por que esta estrategia de entrenamiento permite obtener redes neuronales con mejores capacidades de generalización cuando los datos de entrenamiento estas contaminados con ruido. Esto es concordante con lo observado en [Lopez 07]

Para continuar este trabajo se pretende desarrollar un estudio comparativo de los neuro-controladores inversos entrenados con los algoritmos mencionados en el artículo para generar, de esta manera, conclusiones más generales

Otra posibilidad para complementar este trabajo es usar otros algoritmos de aprendizaje diferentes a los presentados en el documento como por ejemplo las técnicas de entrenamiento basadas en gradiente conjugado